\pdfoutput=1

\documentclass[11pt]{article}

\usepackage[preprint]{acl}

\usepackage{times}
\usepackage{latexsym}

\usepackage[T1]{fontenc}

\usepackage[utf8]{inputenc}

\usepackage{microtype}

\usepackage{inconsolata}

\usepackage{graphicx}
\usepackage{amsmath}
\usepackage{amssymb}
\usepackage{booktabs}
\usepackage{multirow}
\usepackage{colortbl}
\usepackage[normalem]{ulem}
\useunder{\uline}{\ul}{}
\usepackage{float} 
\title{NLoRA: Nyström-Initiated Low-Rank Adaptation for Large Language Models}

\author{
        Chenlu Guo$^{1}$ \quad  Yi Chang$^{1,3,4}$ \quad Yuan Wu$^{1}$\footnotemark[1] \\ 
        $^{1}$School of Artificial Intelligence, Jilin University \\
        $^{2}$Engineering Research Center of Knowledge-Driven Human-Machine Intelligence, MOE, China \\
        $^{3}$International Center of Future Science, Jilin University\\
        guocl23@mails.jlu.edu.cn, yichang@jlu.edu.cn, yuanwu@jlu.edu.cn \\ 
}

\begin{document}
\maketitle

\renewcommand{\thefootnote}{\fnsymbol{footnote}}
\footnotetext[1]{Corresponding authors}

\begin{abstract}
Parameter-efficient fine-tuning (PEFT) is essential for adapting large language models (LLMs), with low rank adaptation (LoRA) being the most popular approach. However, LoRA suffers from slow convergence, and some recent LoRA variants, such as PiSSA, primarily rely on Singular Value Decomposition (SVD) for initialization, leading to expensive computation. To mitigate these problems, we resort to Nyström method, which follows a three-matrix manipulation. Therefore, we first introduce \textbf{S}tructured\textbf{LoRA} (SLoRA), investigating to introduce a small intermediate matrix between the low-rank matrices \(A\) and \(B\). Secondly, we propose \textbf{N}yström\textbf{LoRA} (NLoRA), which leverages Nyström-based initialization for SLoRA to improve its effectiveness and efficiency. Finally, we propose \textbf{Int}ermediate\textbf{Tune} (IntTune) to explore fine-tuning exclusively the intermediate matrix of NLoRA to furthermore boost LLMs' efficiency. 
We evaluate our methods on 5 natural language generation (NLG) tasks and 8 natural language understanding (NLU) tasks. On GSM8K, SLoRA and NLoRA achieve accuracies of 56.48\% and 57.70\%, surpassing LoRA by 33.52\% and 36.41\% with only 3.67M additional trainable parameters. IntTune boosts average NLG performance over LoRA by 7.45\% while using only 1.25\% of its parameters. These results demonstrate the efficiency and effectiveness of our approach in enhancing model performance with minimal parameter overhead.
The code is available at \url{https://github.com/TracyGuo2001/NLoRA}.


\end{abstract}
\section{Introduction}

\begin{figure}[th]
\centering
\includegraphics[width=0.45\textwidth]{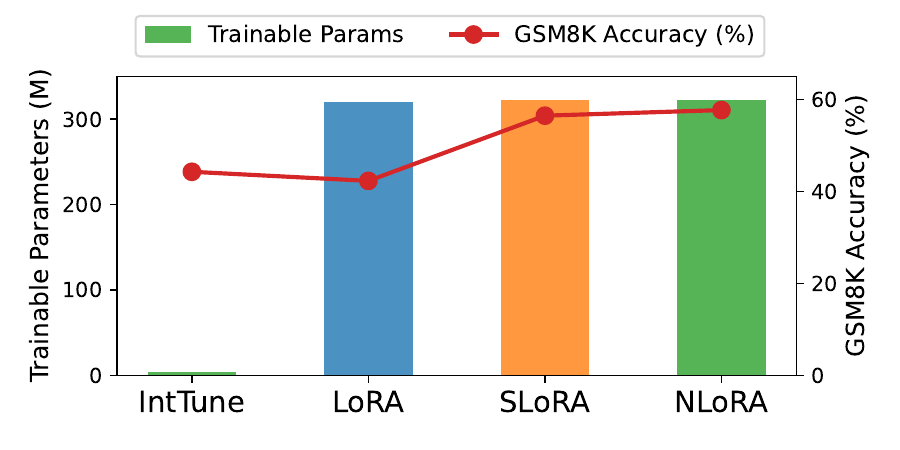} 
\caption{The comparison among LoRA and our models}
\label{figure:compareall}
\end{figure}

\begin{figure*}[th]
\centering
\includegraphics[width=0.9\textwidth]{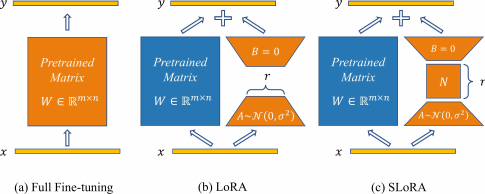} 
\caption{The comparison among Full Fine-tuning, LoRA, and SLoRA}
\label{figure:structure}
\end{figure*}
Fine-tuning large language models (LLMs) has emerged as a fundamental approach to enhancing model capabilities \citep{yu2023metamath,li2023starcoder,xia2024rethinking} and aligning models with specific application requirements \citep{zheng2023judging,ouyang2022training}. However, the growing scale of LLMs introduces significant challenges to LLM development, with fine-tuning requiring substantial computational and memory resources~\citep{hu2021lora,chang2024ba}. For example, fine-tuning a LLaMA-65B model requires more than 780 GB of GPU memory \citep{dettmers2023qlora}, while training GPT-3 175B requires 1.2 TB of VRAM \citep{hu2021lora}. Such resource-intensive processes are infeasible for many researchers and institutions, driving the development of parameter-efficient fine-tuning (PEFT) methods. Among these methods, Low-Rank Adaptation (LoRA) \citep{hu2021lora} has received widespread attention due to its ability to achieve competitive performance compared to full parameter fine-tuning, while significantly reducing memory consumption and avoiding additional inference latency.

LoRA enables the indirect training of dense layers in a neural network by optimizing low-rank decomposition matrices that represent changes in the dense layers during adaptation, all while keeping the pre-trained weights fixed.
For a pre-trained weight matrix \( W \in \mathbb{R}^{m \times n} \), LoRA introduces a low-rank decomposition \( \Delta W = AB \), where \( A \in \mathbb{R}^{m \times r} \), \( B \in \mathbb{R}^{r \times n} \), and the rank \( r \ll \min(m, n) \). This modifies the forward pass of a layer as follows:
\begin{equation}
Y = X(W + \Delta W) = X(W + AB),
\end{equation}
where \( X \in \mathbb{R}^{b \times m} \), \( Y \in \mathbb{R}^{b \times n} \), and \( b \) represents the batch size. For initialization, \( A \) is randomly initialized with Gaussian values and \( B \) is set to zero, ensuring that injection of the low-rank adaptation does not alter the model predictions at the start of training.
Unlike traditional fine-tuning methods that require updating and storing gradients for the full weight matrix \( W \), LoRA optimizes only the smaller matrices \( A \) and \( B \), significantly reducing the number of trainable parameters and memory usage. Furthermore, LoRA often achieves performance comparable or superior to full fine-tuning, demonstrating that adapting only a small subset of parameters suffices for many downstream tasks.

Despite the above benefits, LoRA suffers from slow convergence \cite{ding2023parameter}. To address this issue, some recent LoRA variants, such as PiSSA~\citep{meng2024pissa}, choose to conduct initialization of the low rank matrices by using Singular Value Decomposition (SVD). However, SVD-based initialization is computationally expensive and requires a long time. To mitigate this issue, we investigate using Nyström method, which approximates a matrix as a product of three matrices, to approximate SVD.
To fit the three-matrix structure, we first propose \textbf{S}tructured\textbf{LoRA} (SLoRA), where an additional \(r \times r\) matrix is inserted between the low-rank matrices \(A\) and \(B\), as shown in \figurename~\ref{figure:structure}. 
Furthermore, we explore whether an extra matrix can influence the language model's performance, experimental results indicate that SLoRA effectively enhances performance with only a minor increase in the number of parameters, demonstrating the potential of the three-matrix structure for PEFT.

Secondly, inspired by NyströmFormer~\citep{xiong2021nystromformer}, we proposed \textbf{N}yström\textbf{LoRA} (NLoRA) to leverage Nyström method, which conducts SVD approximation by sampling a subset of rows and columns of the pre-trained parameter matrix to reduce the computational cost, for weight initialization. NLoRA is supposed to bypass the computational cost of SVD's eigenvalue decomposition, reducing time complexity to \(O(mr + r^2 + rn)\) compared to the \(O(mn^2)\) complexity of SVD-based methods.

Finally, to explore whether we can further compress the trainable parameters of NLoRA, we propose \textbf{Int}ermediate\textbf{Tune} (IntTune), which exclusively adjusts the intermediate matrix of NLoRA. This method significantly reduces the number of trainable parameters. Specifically, on the evaluation of LLaMA 2-7B across five NLG benchmarks, LoRA uses 320M parameters, while our IntTune method only requires tuning 4M parameters. In the meantime, IntTune outperforms LoRA by 7.45\% on average across NLG benchmarks. The comparison of our proposed methods with LoRA in terms of performance and trainable parameters is illustrated in \figurename~\ref{figure:compareall}.

In summary, our contributions are as follows:  
\begin{enumerate}  
    \item We propose SLoRA, an extension to the LoRA framework, incorporating an additional intermediate matrix to enhance model expressiveness, achieving improved performance with minimal parameter overhead.  

    \item We introduce NLoRA, leveraging Nyström approximation for efficient and effective initialization, particularly excelling in natural language generation (NLG) and natural language understanding (NLU) tasks.  

    \item We propose IntTune to fulfill supervised fine-tuning (SFT) LLaMA 2-7B by tuning 4M parameters, achieving superior performance compared to LoRA on average, offering a lightweight and efficient alternative for SFT LLMs in resource-constrained scenarios.  
  
\end{enumerate}

\section{Related Works}

\subsection{LoRA's variants} 
With the introduction of LoRA \citep{hu2021lora}, many derivative methods have emerged. 
AdaLORA \citep{zhang2023adalora} highlights that LoRA ignores the importance of different layer parameters based on a uniform setting of the rank, and proposes an adaptive allocation strategy based on parameter importance to improve fine-tuning efficiency.
DoRA \citep{liu2024dora} introduces a decomposation of weight matrices into magnitude and direction components, leveraging LoRA to update only the directional component, thereby reducing the number of trainable parameters. ReLoRA \citep{lialin2023relora} achieves high-rank training through iterative low-rank updates, periodically merging parameters into the main model. LoRA+ \citep{hayou2024lora+} further improves efficiency by applying different learning rates to the two matrices in LoRA, assigning a higher learning rate to matrix \(B\) to accelerate convergence and enhance performance.   Other works have focused on improving the initialization of the \(AB\) matrix, such as PiSSA \citep{meng2024pissa}, which suggests initializing \(A\) and \(B\) by performing SVD on the pre-trained matrix \(W\) to accelerate the convergence speed. LoRA-GA \citep{wang2024lora} initializes \(A\) and \(B\) using the eigenvectors of the full-gradient matrix, aligning the gradient direction of the low-rank product \(BA\) with the gradient direction of the pretrained weight matrix \(W\).
\begin{figure*}[]
\centering
\includegraphics[width=0.9\textwidth]{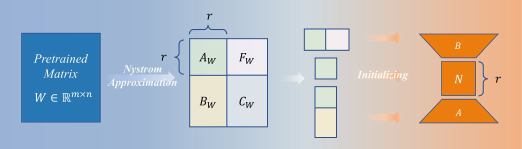} 
\caption{The diagram of the Nyström-based initialization}
\label{figure:nlora}
\end{figure*}
\subsection{Nyström-like methods}
Nyström-like methods approximate matrices by sampling a subset of columns, a technique widely used in kernel matrix approximation \citep{baker1979numerical,williams2000using}. Numerous variants have been proposed to enhance the basic Nyström method, including Nyström with k-means clustering \citep{wang2019scalable}, Nyström with spectral problems \citep{vladymyrov2016variational}, randomized Nyström \citep{li2010making,persson2024randomized}, ensemble Nyström method \citep{kumar2009ensemble}, fast-Nys \citep{si2016computationally}.

The Nyström method has also been extended to general matrix approximation beyond symmetric matrices \citep{nemtsov2016matrix}. Some methods \citep{wang2013improving,xiong2021nystromformer} explicitly address general matrix approximation by sampling both rows and columns to reconstruct the full matrix. Inspired by such strategies, we propose NLoRA method by to optimize the approximation for efficient matrix reconstruction.

\section{Method}
\label{sec:method}
The Nyström method \cite{baker1979numerical}, originating from the field of integral equations, is a approach for discretizing integral equations using a quadrature technique. It is commonly employed for out-of-sample extension problems. Specifically, given an eigenfunction problem of the form:
\begin{equation}
\lambda f(x) = \int_a^b M(x, y) f(y) \, dy,
\end{equation}
the Nyström method utilizes a set of \( s \) sample points \( y_1, y_2, \ldots, y_s \) to approximate \( f(x) \) as follows:
\begin{equation}
\lambda \tilde{f}(x) \triangleq \frac{b-a}{s} \sum_{j=1}^{s} M(x, y_j) f(y_j).
\end{equation}
This approach effectively converts the continuous integral equation into a discrete summation, facilitating numerical computation and enabling out-of-sample extensions.

For the pre-trained matrix \( W \in \mathbb{R}^{m \times n} \), we assume that it can be decomposed as follows:
\begin{equation}
W = \begin{bmatrix}
A_W & B_W \\
F_W & C_W
\end{bmatrix},  \label{eq:Wsplit}
\end{equation}
where, \( A_W \in \mathbb{R}^{r \times r} \) is designated to be our sample matrix, \( B_W \in \mathbb{R}^{r \times (n-r)} \) and \( F_W \in \mathbb{R}^{(m-r) \times r} \) represent the remaining sampled column and row components, respectively, and \( C_W \in \mathbb{R}^{(m-r) \times (n-r)} \) corresponds to the remainder of the matrix \( W \).
The matrix \( W \) can be efficiently approximated using the Nyström method's basic quadrature technique. Starting with the singular value decomposition (SVD) of the sample matrix \( A_W \), represented as \( A_W = U \Lambda V^T \), where \( U, V \in \mathbb{R}^{r \times r} \) are unitary matrices and \( \Lambda \in \mathbb{R}^{r \times r} \) is diagonal. 
The Nyström approximation reconstructs \( W \) based on the out-of-sample approximation strategy \citep{nemtsov2016matrix}. 
This strategy utilizes the entries of \( F_{W} \) and \( B_{W} \) as interpolation weights for extending the singular vector, resulting in the full approximations of the left and right singular vectors of \( W \):
\begin{equation}
\hat{U} = 
\begin{bmatrix}
U \\
F_{W} V \Lambda^{-1}
\end{bmatrix},
\quad
\hat{V} = 
\begin{bmatrix}
V \\
B_{W}^{T} U \Lambda^{-1}
\end{bmatrix},
\end{equation}

Using the Nyström method, the pretrained matrix \( W \) can be approximated as:
\begin{align}
\widehat{W} &= \hat{U}\Lambda \hat{V}^{T} =
\begin{bmatrix}
A_{W} & B_{W} \\
F_{W} & F_{W} A_{W}^+ B_{W}
\end{bmatrix} \nonumber\\
&=
\begin{bmatrix}
A_{W}\\
F_{W}
\end{bmatrix} A_{W}^+ 
\begin{bmatrix}
A_{W} & B_{W}
\end{bmatrix}, 
\end{align}
where \( A_{W}^+ \) is the Moore-Penrose pseudoinverse of the sampled core matrix \( A_{W} \). The remaining block \(C_{W} \) is approximated as \( F_{W} A_{W}^+ B_{W} \). This approximation demonstrates that \( W \) can be effectively reconstructed using only \( A_{W} \), \( B_{W} \), and \( F_{W} \), significantly reducing computational complexity. For the detailed derivation, please refer to \appendixname~\ref{app:derivation}.

In this way, the matrix \( W \) can be approximated as the product of three matrices. Based on this finding, we propose an improvement to LoRA by introducing an intermediate matrix, named as \textbf{S}tructured\textbf{LoRA} (SLoRA). Specifically, we introduce an intermediate matrix \( N \in \mathbb{R}^{r \times r} \) between the low-rank matrices \( A \) and \( B \), as illustrated in \figurename~\ref{figure:structure}. This modification transforms the weight update into:
\begin{equation}
\Delta W = A N B,
\end{equation}
where \( A \in \mathbb{R}^{m \times r} \), \( B \in \mathbb{R}^{r \times n} \), \( N \in \mathbb{R}^{r \times r} \), and \( r \ll \min(m, n) \). 

Building on the three-matrix structure, we further enhance SLoRA's effectiveness by employing a Nyström-based initialization. Specifically, by sampling \( r \) rows and \( r \) columns—corresponding to the rank of LoRA—we efficiently approximate \( W \) through matrix decomposition. The resulting submatrices are then directly utilized to initialize the three components of SLoRA, specifically:
\begin{itemize}
    \item The component $\begin{bmatrix} A_{W} \\ F_{W} \end{bmatrix}$ is used to initialize the matrix $A$ in SLoRA.
    \item The component $A_{W}^+$, representing the Moore-Penrose pseudoinverse of $A_{W}$, is used to initialize the matrix $N$ in SLoRA.
    \item The component $\begin{bmatrix} A_{W} & B_{W} \end{bmatrix}$ is used to initialize the matrix $B$ in SLoRA.
\end{itemize}
While the pseudoinverse can be computed using singular value decomposition (SVD), the process is computationally inefficient on GPUs. To overcome this challenge, we simplify the initialization by directly employing \( A_{W} \) instead of its pseudoinverse, thereby reducing computational overhead while preserving the effectiveness of the initialization. The diagram of the Nyström-based initialization is shown in \figurename~\ref{figure:nlora}.

\begin{table*}[h]
\centering
\begin{tabular}{cccccccc}
\toprule
\textbf{Model}               & \textbf{Strategy} & \textbf{Parameters} & \textbf{GSM8K}                & \textbf{MATH}                          & \textbf{HumanEval}                    & \textbf{MBPP}                 & \textbf{MT-Bench}            \\ 
\midrule
                             & Full FT           & 6738M               & 49.05                         & 7.22                                   & 21.34                                 & 35.59                         & \textbf{4.91}                \\
                             & LoRA              & 320M                & 42.30                         & 5.50                                   & 18.29                                 & 35.34                         & 4.58                         \\
                             & PiSSA             & 320M                & 53.07                         & 7.44                                   & 21.95                                 & 37.09                         & 4.87                         \\ 
\cmidrule{2-8} 
                             & SLoRA             & 323M                & \cellcolor[HTML]{ECF4FF}56.48 & \cellcolor[HTML]{ECF4FF}\textbf{10.68} & \cellcolor[HTML]{ECF4FF}23.78         & \cellcolor[HTML]{ECF4FF}42.32 & \cellcolor[HTML]{ECF4FF}4.85 \\
\multirow{-5}{*}{LLaMA 2-7B} & NLoRA             & 323M                & \textbf{57.70}                & 9.94                                   & \textbf{25.00}                        & \textbf{43.12}                & 4.82                         \\ 
\midrule
                             & Full FT           & 7242M               & 67.02                         & 18.6                                   & 45.12                                 & 51.38                         & 4.95                         \\
                             & LoRA              & 168M                & 67.70                         & 19.68                                  & 43.90                                 & 58.39                         & 4.90                         \\
                             & PiSSA             & 168M                & 72.86                         & 21.54                                  & 46.95                                 & \textbf{62.66}                & \textbf{5.34}                \\ 
\cmidrule{2-8} 
                             & SLoRA             & 169M                & \cellcolor[HTML]{ECF4FF}73.01 & \cellcolor[HTML]{ECF4FF}21.88          & \cellcolor[HTML]{ECF4FF}\textbf{47.6} & \cellcolor[HTML]{ECF4FF}60.3  & \cellcolor[HTML]{ECF4FF}5.12 \\
\multirow{-5}{*}{Mistral-7B} & NLoRA             & 169M                & \textbf{73.92}                & \textbf{22.00}                         & 44.5                                  & 60.3                          & 5.21                         \\ 
\bottomrule
\end{tabular}
\caption{Experimental results on NLG tasks}
\label{table:nlg}
\end{table*}
\begin{table*}[h]
\centering
\begin{tabular}{ccccccccc}
\toprule
\textbf{Strategy} & \textbf{MNLI}                         & \textbf{SST-2}                        & \textbf{MRPC}                          & \textbf{CoLA}                 & \textbf{QNLI}                 & \textbf{QQP}                  & \textbf{RTE}                           & \textbf{STS-B}                \\ 
\midrule
\multicolumn{9}{c}{\textbf{DeBERTa-v3-base}} \\ 
\midrule
Full FT           & 89.90                                 & 95.63                                 & 89.46                                  & 69.19                         & 94.03                         & \textbf{92.40}                & 83.75                                  & 91.60                         \\
LoRA              & 90.65                                 & 94.95                                 & 89.95                                  & 69.82                         & 93.87                         & 91.99                         & 85.20                                  & 91.60                         \\
PiSSA             & 90.43                                 & 95.87                                 & 91.67                                  & 72.64                         & 94.29                         & 92.26                         & 87.00                                  & 91.88                         \\
SLoRA             & 90.43                                 & \cellcolor[HTML]{ECF4FF}96.10         & \cellcolor[HTML]{ECF4FF}\textbf{91.91} & \cellcolor[HTML]{ECF4FF}70.82 & \cellcolor[HTML]{ECF4FF}93.94 & \cellcolor[HTML]{ECF4FF}92.11 & \cellcolor[HTML]{ECF4FF}\textbf{88.09} & \cellcolor[HTML]{ECF4FF}91.86 \\
NLoRA             & \textbf{90.74}                        & \textbf{96.22}                        & \textbf{91.91}                         & \textbf{73.41}                & \textbf{94.45}                & 92.03                         & \textbf{88.09}                         & \textbf{92.14}                \\ 
\midrule
\multicolumn{9}{c}{\textbf{RoBERTa-large}} \\ 
\midrule
Full FT           & 90.2                                  & 96.4                                  & 90.9                                   & 68.0                          & 94.7                          & \textbf{92.2}                 & 86.6                                   & 91.5                          \\
LoRA              & 90.6                                  & 96.2                                  & 90.9                                   & 68.2                          & 94.9                          & 91.6                          & 87.4                                   & 92.6                          \\
PiSSA             & 90.7                                  & 96.7                                  & \textbf{91.9}                                   & 69.0                          & 95.1                          & 91.6                          & \textbf{91.0}                          & \textbf{92.9}                 \\
SLoRA             & \cellcolor[HTML]{ECF4FF}\textbf{90.8} & \cellcolor[HTML]{ECF4FF}\textbf{96.8} & \cellcolor[HTML]{ECF4FF}91.7           & \cellcolor[HTML]{ECF4FF}68.5  & \cellcolor[HTML]{ECF4FF}94.9  & \cellcolor[HTML]{ECF4FF}91.6  & \cellcolor[HTML]{ECF4FF}90.3           & \cellcolor[HTML]{ECF4FF}92.7  \\
NLoRA             & 90.7                                  & 96.6                                  & \textbf{91.9}                          & \textbf{69.7}                 & \textbf{95.2}                 & 91.6                          & 90.3                                   & 92.7                          \\ 
\bottomrule
\end{tabular}
\caption{Experimental results on NLU tasks}
\label{table:nlu}
\end{table*}

By employing this decomposition based on the Nyström approximation method, we propose an initialization strategy for SLoRA, which we term as \textbf{N}yström\textbf{LoRA} (NLoRA). Additionally, we explore fine-tuning only the intermediate matrix while keeping the other two matrices fixed,  which we term \textbf{Int}ermediate\textbf{Tune} (IntTune). 

\section{Experiments}
\label{sec:experiments}
The experiments were performed on NVIDIA L20 GPUs. For these experiments, we follow the experimental setting given by~\cite{meng2024pissa}, we employ the AdamW optimizer with a batch size of 4, a learning rate of 2E-4, and a cosine annealing schedule with a warmup ratio of 0.03, all while avoiding weight decay. The parameter lora\_alpha is consistently set equal to lora\_r, with lora\_dropout fixed at 0. Adapters are integrated into all linear layers of the base model, and both the base model and adapters utilized Float32 precision for computation. We take the convenience to directly cite the baseline performance values from~\cite{meng2024pissa}.

In this section, we evaluate the performance of SLoRA and NLoRA across various benchmark datasets. 
We compare them with the following baselines:
\begin{itemize}
    \item Full Fine-tune: which updates all model parameters;
    \item LoRA \cite{hu2021lora}: which approximates weight updates with low-rank matrices while freezing the base model;
    \item PiSSA \cite{meng2024pissa}: which initializes adapters using principal singular components and freezes residuals while retaining LoRA's architecture.
\end{itemize}

We evaluate the capabilities of natural language generation (NLG) using the LLaMA 2-7B \citep{touvron2023llama} and Mistral-7B \citep{jiang2023mistral} models through mathematical reasoning, coding proficiency, and dialogue tasks. Additionally, natural language understanding (NLU) tasks were evaluated using the GLUE dataset \citep{wang2018glue} with DeBERTa-v3-base \citep{he2021debertav3} and RoBERTa-large \citep{liu2019roberta}. Finally, we analyze the empirical effects of exclusively fine-tuning the intermediate matrix on both NLU and NLG tasks.
\begin{table*}[t]
\centering
\begin{tabular}{@{}ccccccc@{}}
\toprule
\textbf{Strategy} & \textbf{Parameters} & \textbf{GSM8K} & \textbf{MATH} & \textbf{HumanEval} & \textbf{MBPP} & \textbf{MT-Bench} \\ \midrule
LoRA              & 320M                & 42.30          & 5.50          & 18.29              & 35.34         & 4.58              \\
IntTune           & 4M                  & 44.28          & 6.86          & 20.70               & 34.40          & 4.46              \\ \bottomrule
\end{tabular}
\caption{IntTune performance on NLG tasks}
\label{table:llama_on_results}
\end{table*}

\begin{table*}[t]
\centering
\begin{tabular}{@{}cccccccccc@{}}
\toprule
\textbf{Strategy} & \textbf{Parameters} & \textbf{MNLI} & \textbf{SST-2} & \textbf{MRPC} & \textbf{CoLA} & \textbf{QNLI} & \textbf{QQP} & \textbf{RTE} & \textbf{STS-B} \\ \midrule
LoRA              & 1.33M               & 90.65         & 94.95          & 89.95         & 69.82         & 93.87         & 91.99        & 85.20        & 91.60          \\
IntTune           & 3.07K              & 81.93         & 92.20          & 85.29         & 65.38         & 89.13         & 85.18        & 76.90        & 88.37          \\ \bottomrule
\end{tabular}
\caption{IntTune performance on NLU tasks}
\label{table:deberta_on_results}
\end{table*}

\subsection{Experiments on Natural Language Generation}
\label{sec:nlg}
We conduct experiments using LLaMA 2-7B and Mistral-7B-v0.1. To evaluate mathematical reasoning abilities, we perform fine-tuning using the MetaMathQA dataset and evaluated their performance on GSM8K \citep{cobbe2021training} and MATH \citep{yu2023metamath}. In terms of coding capability, we perform fine-tuning on the CodeFeedback dataset \citep{zheng2024opencodeinterpreter} and evaluated them using the HumanEval \citep{chen2021evaluating} and MBPP \citep{austin2021program} benchmarks. To measure session capabilities, the model is fine-tuned on the WizardLM-Evol-Instruct dataset \citep{xu2024wizardlm} and tested using the MT-Bench dataset \citep{zheng2023judging}. All experiments use a subset of 100K data points.

As shown in \tablename~\ref{table:nlg}, SLoRA consistently outperforms LoRA, which is labeled with a blue background in \tablename~\ref{table:nlg}, and even outperforms PiSSA in most tasks. In most cases, NLoRA further enhances the performance of SLoRA. 
Both methods maintain high parameter efficiency, with only slight increases in trainable parameters (1.15\% for LLaMA 2-7B and 0.55\% for Mistral-7B compared to LoRA), yet deliver significant performance gains. 
On these two models, SLoRA achieves average improvements of 38.68\%, 15.37\%, and 5.19\% in mathematical reasoning, coding, and conversational tasks, respectively, relative to LoRA's performance, while NLoRA achieves improvements of 34.53\%, 15.83\%, and 5.78\% over LoRA.

Although the addition of intermediate matrices results in additional matrix multiplication operations, the time overhead increases only slightly compared to LoRA. In the MetaMathQA dataset, the training time for SLoRA increases to 27,690.03 seconds, which is an increase of 10. 13\% compared to LoRA (25142.26 seconds). The training time for NLoRA increases to 25,323.34 seconds, which is almost identical to LoRA's training time.
As for initialization time, as shown in \tablename~ \ref{tab:init_time}, SLoRA incurs only an 11.95\% increase in initialization time compared to LoRA, while NLoRA adds just 12.66 seconds. Both are significantly lower than the time cost of PiSSA.
\begin{table}[]
\centering
\begin{tabular}{@{}cc@{}}
\toprule
\textbf{Strategy} & \textbf{Time (seconds)} \\ \midrule
SLoRA           & 14.21                                  \\
NLoRA           & 25.35                                \\
LoRA            & 12.69                                  \\
PiSSA           & 106903.20                              \\ \bottomrule
\end{tabular}
\caption{Initialization time of different strategies}
\label{tab:init_time}
\end{table}
Subsequently, we further discuss the effects under different ranks (Section \ref{app:r}), learning rates (\appendixname~\ref{app:lr}), and optimizers (\appendixname~\ref{app:optim}).

\begin{figure*}[th]
\centering
\includegraphics[width=1\textwidth]{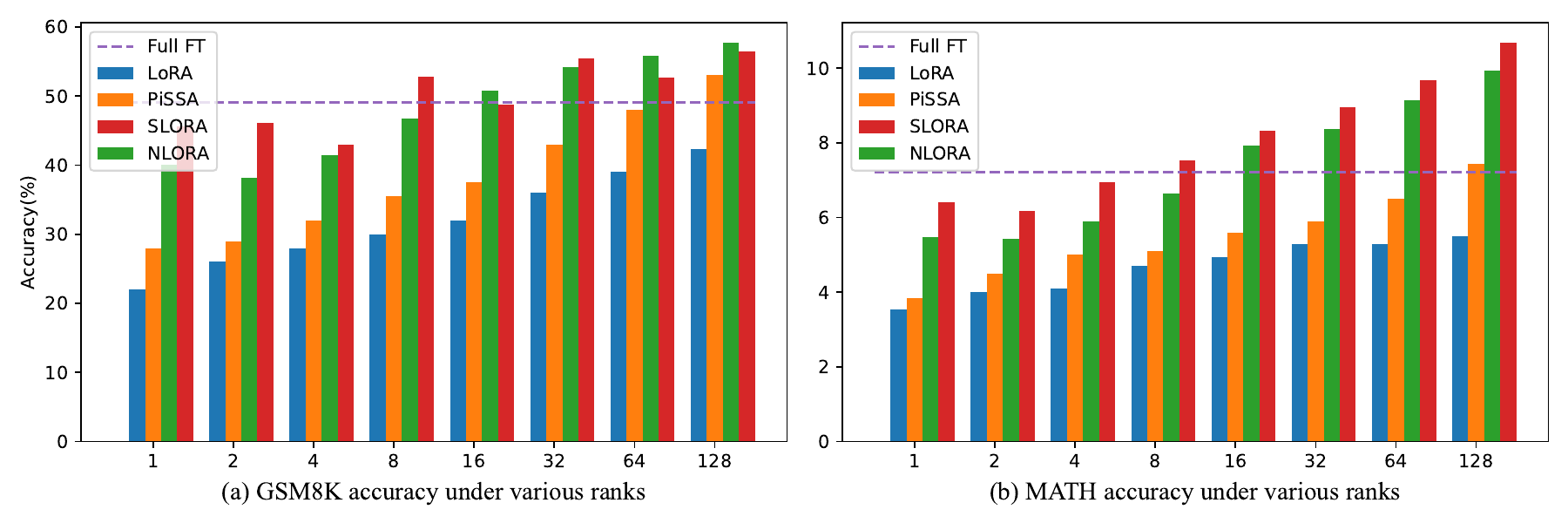} 
\caption{Compare the performance of different ranks for NLoRA on NLG tasks}
\label{figure:rank_comparison}
\end{figure*}

\subsection{Experiments on Natural Language Understanding}
We also assess the NLU capabilities of RoBERTa-large and DeBERTa-v3-base on the GLUE benchmark. \tablename~\ref{table:nlu} summarizes the results of eight tasks performed using these two base models. 

SLoRA demonstrates consistent improvements over the baseline LoRA across all tasks, as highlighted in blue. In addition, SLoRA surpasses PiSSA in several cases, showcasing the potential of incorporating an intermediate matrix in LoRA. NLoRA further enhances the performance of SLoRA in most tasks, achieving superior results in tasks such as QNLI, MRPC, and CoLA. For instances where NLoRA does not outperform PiSSA, NLoRA consistently achieves a lower training loss in these scenarios, suggesting its potential for further optimization and efficient fine-tuning. Details can be found in \appendixname~\ref{app:nlu}.

\subsection{NLoRA's Intermediate Matrix Fine-Tuning: A Minimalist Approach}

To further improve the computational efficiency of NLoRA, we try to investigate reducing its trainable parameters without sacrificing much performance. Therefore, we propose \textbf{Int}ermediate\textbf{Tune} (IntTune), which exclusively fine-tune the intermediate matrix in SFT. To validate the effectiveness of IntTune, we conduct experiments using LLaMA-2-7B and DeBERTa-v3-base for NLG and NLU tasks, respectively.
For NLG tasks, we set the learning rate to 2E-3 while keeping other settings unchanged. For NLU tasks, the specific parameter settings are detailed in \appendixname~\ref{app:nlu}. 
The results are shown in Table~\ref{table:llama_on_results} and Table~\ref{table:deberta_on_results}.

\begin{figure}[]
\centering
\includegraphics[width=0.5\textwidth]{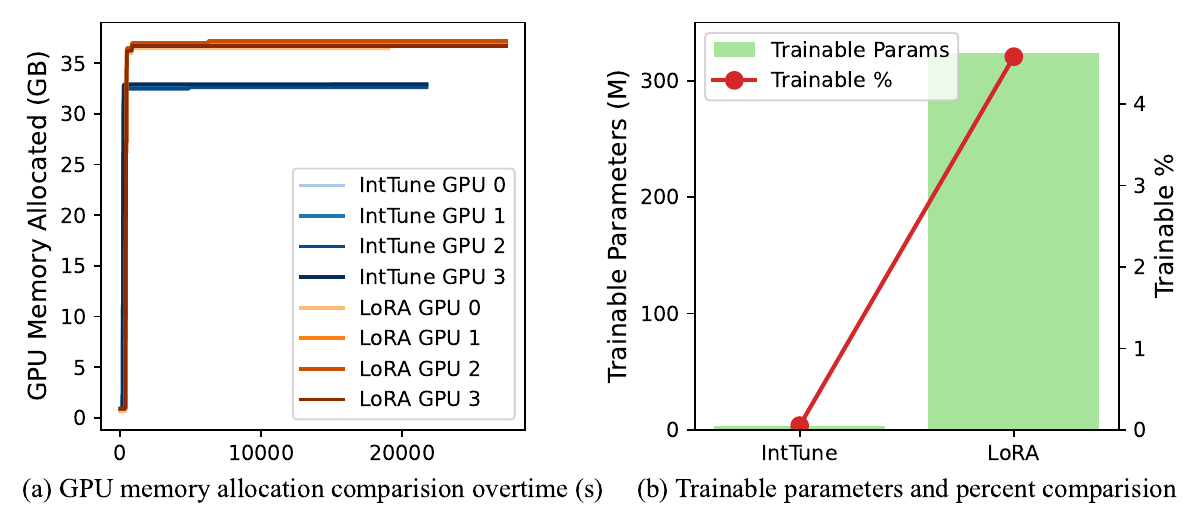} 
\caption{Comparison of GPU memory allocation and trainable parameters between IntTune and LoRA}
\label{figure:on128}
\end{figure}

\begin{table*}[]
\centering
\begin{tabular}{@{}ccccccc@{}}
\toprule
\textbf{Strategy} & \textbf{Parameters} & \textbf{GSM8K} & \textbf{MATH} & \textbf{HumanEval} & \textbf{MBPP} & \textbf{MT-Bench} \\ \midrule
LoRA              & 320M                & 42.30          & 5.50          & 18.29              & 35.34         & \textbf{4.58}              \\
IntTune(Rank=256) & 15M                 & \textbf{49.51}          & 6.62          & \textbf{21.30}              & 33.90         & 3.59              \\
IntTune(Rank=128) & 4M                  & 44.28          & \textbf{6.86}          & 20.70              & 34.40         & 4.46              \\
IntTune(Rank=64)  & 0.9M                & 37.98          & 5.56          & 14.60              & \textbf{34.70}         & 4.55              \\ \bottomrule
\end{tabular}
\caption{Compare the performance of different ranks for IntTune on NLG tasks}
\label{table:rank_on_comparisonNLG}
\end{table*}

\begin{figure*}[t]
\centering
\includegraphics[width=0.95\textwidth]{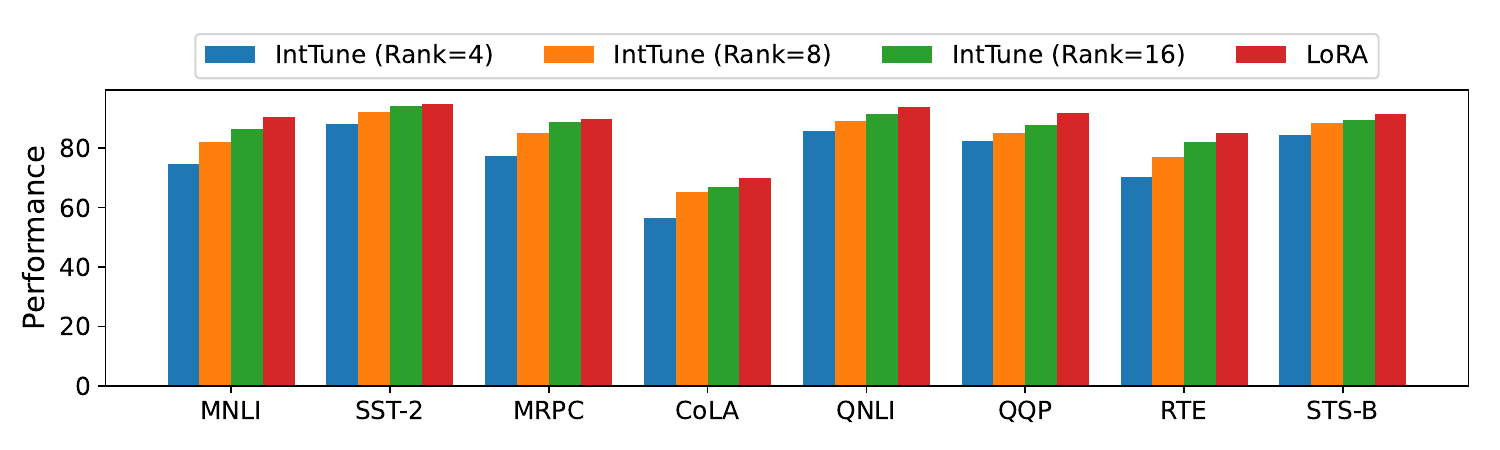} 
\caption{Compare the performance of different ranks for IntTune on NLU tasks}
\label{figure:rank_on_comparisonNLU}
\end{figure*}

For NLG tasks, IntTune achieves competitive performance, surpassing LoRA on the GSM8K, MATH, and HumanEval tasks, and attaining comparable results on MBPP and MT-Bench. Overall, the average performance of IntTune across all tasks exceeds that of LoRA, surpassing LoRA's average performance by 7.45\%. 
The comparison of training parameters and memory allocation between IntTune and LoRA is shown in \figurename~\ref{figure:on128}, with all measurements recorded on the MetaMathQA dataset.
In terms of computational efficiency, IntTune significantly reduces the number of trainable parameters to 4M, accounting for only 0.05\% of the total model parameters and just 1.13\% of LoRA's trainable parameters.
Despite this substantial reduction, the training time is shortened to 85.2\% of LoRA's. Specifically, LoRA's training time is 25,142.27s, IntTune's training time is reduced to 21,439.26s. Additionally, IntTune enables GPU memory allocation to decrease as well. The percentage of GPU memory allocated drops from 80.9\% to 72.5\%, with the average memory usage reduced from 36.42 GB to 32.78 GB, a reduction of 9.98\%. These results highlight the method's potential for improving performance while optimizing computational resources, making it particularly suitable for SFT LLMs in resource-constrained scenarios.

For NLU tasks, the number of trainable parameters was reduced to 3.07K, representing 0.002\% of the total model parameters. Despite this significant reduction, the approach achieved 92.61\% of LoRA's average performance across all tasks. Specifically, it attained 96.2\% of LoRA's performance on SST-2, 94.5\% on QNLI, and 96.2\% on STS-B, demonstrating comparable performance across various GLUE tasks, underscoring its robustness and effectiveness in diverse scenarios.

These results demonstrate the potential of the Nyström initialization, as fine-tuning only the intermediate matrix can still yield competitvie performance.

\subsection{Experiments on Various Ranks}
\label{app:r}
In this section, we examine the impact of progressively increasing the rank of NLoRA and SLoRA from 1 to 128 to assess their ability to consistently outperform the baseline across different ranks. Training is performed on the MetaMathQA dataset for a single epoch, with validation conducted on the GSM8K and MATH datasets. 

The experimental results are presented in \figurename~\ref{figure:rank_comparison}.
On the GSM8K dataset, NLoRA performs relatively better at higher ranks, surpassing LoRA by 43.08\% and 36.41\% at ranks 64 and 128, respectively. SLoRA, on the other hand, exhibits relatively stronger performance at lower ranks, outperforming LoRA by 107.45\%, 77.31\%, 53.54\%, and 76.13\% at ranks 1, 2, 4, and 8, respectively.
On the MATH dataset, SLoRA shows a slight overall advantage, while NLoRA continues to deliver strong performance, particularly at higher ranks.

For IntTune, we compared ranks of 64, 128, and 256 in the NLG tasks, following the same experimental setup as shown in Section 4.1. In the NLU experiments, we evaluated ranks of 4, 8, and 16. The results of these experiments are presented in Table~\ref{table:rank_on_comparisonNLG} and \figurename~\ref{figure:rank_on_comparisonNLU}.
On NLG tasks, IntTune does not exhibit a strictly increasing performance trend with higher ranks. Instead, different ranks excel in different tasks. 
Specifically, rank 128 and rank 256 achieve 7.45\% and 5.62\% higher performance than LoRA on average, both outperforming LoRA overall.
Meanwhile, rank 64, though slightly lower, still reaches 93.66\% of LoRA’s performance, demonstrating the feasibility of fine-tuning with even fewer parameters while maintaining competitive results.
On NLU tasks, the model performance gradually improves with increasing rank. For ranks 4, 8, and 16, the average performance reaches 86.20\%, 92.61\%, and 95.80\% of LoRA's performance, respectively, while the number of parameters is only 1.35K, 3.07K, and 9.99K, respectively. 


\section{Conclusion}
This work advances parameter-efficient fine-tuning strategies for large language models by introducing SLoRA and NLoRA, along with an exploration of an intermediate matrix fine-tuning method, IntTune. SLoRA incorporates a small intermediate matrix, enhancing expressiveness with minimal parameter overhead, while NLoRA leverages Nyström-based initialization to bypass the computational complexity of SVD, achieving competitive downstream performance. 
IntTune, by fine-tuning only the intermediate matrix in NLoRA, even  boosts average NLG performance over LoRA while maintaining high parameter efficiency.
Extensive experiments on NLG and NLU tasks demonstrate the robustness and adaptability of our methods, providing practical solutions for optimizing large models under resource constraints. 

\section{Limitaion}
While our method demonstrates strong performance in both NLG and NLU tasks, its applicability to ultra-low parameter fine-tuning approaches, such as IntTune, warrants further exploration. Additionally, extending our approach to visual tasks could provide valuable insights into its generalization and versatility across modalities. 
Furthermore, integrating SLoRA with advanced LoRA variants presents a compelling direction for future research to further enhance fine-tuning efficacy.

\bibliography{custom}

\begin{thebibliography}{43}
\providecommand{\natexlab}[1]{#1}

\bibitem[{Aho and Ullman(1972)}]{Aho:72}
Alfred~V. Aho and Jeffrey~D. Ullman. 1972.
\newblock \emph{The Theory of Parsing, Translation and Compiling}, volume~1.
\newblock Prentice-Hall, Englewood Cliffs, NJ.

\bibitem[{{American Psychological Association}(1983)}]{APA:83}
{American Psychological Association}. 1983.
\newblock \emph{Publications Manual}.
\newblock American Psychological Association, Washington, DC.

\bibitem[{Ando and Zhang(2005)}]{Ando2005}
Rie~Kubota Ando and Tong Zhang. 2005.
\newblock A framework for learning predictive structures from multiple tasks and unlabeled data.
\newblock \emph{Journal of Machine Learning Research}, 6:1817--1853.

\bibitem[{Andrew and Gao(2007)}]{andrew2007scalable}
Galen Andrew and Jianfeng Gao. 2007.
\newblock Scalable training of {L1}-regularized log-linear models.
\newblock In \emph{Proceedings of the 24th International Conference on Machine Learning}, pages 33--40.

\bibitem[{Austin et~al.(2021)Austin, Odena, Nye, Bosma, Michalewski, Dohan, Jiang, Cai, Terry, Le et~al.}]{austin2021program}
Jacob Austin, Augustus Odena, Maxwell Nye, Maarten Bosma, Henryk Michalewski, David Dohan, Ellen Jiang, Carrie Cai, Michael Terry, Quoc Le, et~al. 2021.
\newblock Program synthesis with large language models.
\newblock \emph{arXiv preprint arXiv:2108.07732}.

\bibitem[{Baker and Taylor(1979)}]{baker1979numerical}
Christopher~TH Baker and RL~Taylor. 1979.
\newblock The numerical treatment of integral equations.
\newblock \emph{Journal of Applied Mechanics}, 46(4):969.

\bibitem[{Chandra et~al.(1981)Chandra, Kozen, and Stockmeyer}]{Chandra:81}
Ashok~K. Chandra, Dexter~C. Kozen, and Larry~J. Stockmeyer. 1981.
\newblock \href {https://doi.org/10.1145/322234.322243} {Alternation}.
\newblock \emph{Journal of the Association for Computing Machinery}, 28(1):114--133.

\bibitem[{Chang et~al.(2024)Chang, Chang, and Wu}]{chang2024ba}
Yupeng Chang, Yi~Chang, and Yuan Wu. 2024.
\newblock Ba-lora: Bias-alleviating low-rank adaptation to mitigate catastrophic inheritance in large language models.
\newblock \emph{arXiv preprint arXiv:2408.04556}.

\bibitem[{Chen et~al.(2021)Chen, Tworek, Jun, Yuan, Pinto, Kaplan, Edwards, Burda, Joseph, Brockman et~al.}]{chen2021evaluating}
Mark Chen, Jerry Tworek, Heewoo Jun, Qiming Yuan, Henrique Ponde De~Oliveira Pinto, Jared Kaplan, Harri Edwards, Yuri Burda, Nicholas Joseph, Greg Brockman, et~al. 2021.
\newblock Evaluating large language models trained on code.
\newblock \emph{arXiv preprint arXiv:2107.03374}.

\bibitem[{Cobbe et~al.(2021)Cobbe, Kosaraju, Bavarian, Chen, Jun, Kaiser, Plappert, Tworek, Hilton, Nakano et~al.}]{cobbe2021training}
Karl Cobbe, Vineet Kosaraju, Mohammad Bavarian, Mark Chen, Heewoo Jun, Lukasz Kaiser, Matthias Plappert, Jerry Tworek, Jacob Hilton, Reiichiro Nakano, et~al. 2021.
\newblock Training verifiers to solve math word problems.
\newblock \emph{arXiv preprint arXiv:2110.14168}.

\bibitem[{Dettmers et~al.(2023)Dettmers, Pagnoni, Holtzman, and Zettlemoyer}]{dettmers2023qlora}
Tim Dettmers, Artidoro Pagnoni, Ari Holtzman, and Luke Zettlemoyer. 2023.
\newblock Qlora: efficient finetuning of quantized llms (2023).
\newblock \emph{arXiv preprint arXiv:2305.14314}, 52:3982--3992.

\bibitem[{Ding et~al.(2023)Ding, Qin, Yang, Wei, Yang, Su, Hu, Chen, Chan, Chen et~al.}]{ding2023parameter}
Ning Ding, Yujia Qin, Guang Yang, Fuchao Wei, Zonghan Yang, Yusheng Su, Shengding Hu, Yulin Chen, Chi-Min Chan, Weize Chen, et~al. 2023.
\newblock Parameter-efficient fine-tuning of large-scale pre-trained language models.
\newblock \emph{Nature Machine Intelligence}, 5(3):220--235.

\bibitem[{Gusfield(1997)}]{Gusfield:97}
Dan Gusfield. 1997.
\newblock \emph{Algorithms on Strings, Trees and Sequences}.
\newblock Cambridge University Press, Cambridge, UK.

\bibitem[{Hayou et~al.(2024)Hayou, Ghosh, and Yu}]{hayou2024lora+}
Soufiane Hayou, Nikhil Ghosh, and Bin Yu. 2024.
\newblock Lora+: Efficient low rank adaptation of large models.
\newblock \emph{arXiv preprint arXiv:2402.12354}.

\bibitem[{He et~al.(2021)He, Gao, and Chen}]{he2021debertav3}
Pengcheng He, Jianfeng Gao, and Weizhu Chen. 2021.
\newblock Debertav3: Improving deberta using electra-style pre-training with gradient-disentangled embedding sharing.
\newblock \emph{arXiv preprint arXiv:2111.09543}.

\bibitem[{Hu et~al.(2021)Hu, Shen, Wallis, Allen-Zhu, Li, Wang, Wang, and Chen}]{hu2021lora}
Edward~J Hu, Yelong Shen, Phillip Wallis, Zeyuan Allen-Zhu, Yuanzhi Li, Shean Wang, Lu~Wang, and Weizhu Chen. 2021.
\newblock Lora: Low-rank adaptation of large language models.
\newblock \emph{arXiv preprint arXiv:2106.09685}.

\bibitem[{Jiang et~al.(2023)Jiang, Sablayrolles, Mensch, Bamford, Chaplot, Casas, Bressand, Lengyel, Lample, Saulnier et~al.}]{jiang2023mistral}
Albert~Q Jiang, Alexandre Sablayrolles, Arthur Mensch, Chris Bamford, Devendra~Singh Chaplot, Diego de~las Casas, Florian Bressand, Gianna Lengyel, Guillaume Lample, Lucile Saulnier, et~al. 2023.
\newblock Mistral 7b.
\newblock \emph{arXiv preprint arXiv:2310.06825}.

\bibitem[{Kumar et~al.(2009)Kumar, Mohri, and Talwalkar}]{kumar2009ensemble}
Sanjiv Kumar, Mehryar Mohri, and Ameet Talwalkar. 2009.
\newblock Ensemble nystrom method.
\newblock \emph{Advances in Neural Information Processing Systems}, 22.

\bibitem[{Li et~al.(2010)Li, Kwok, and L{\"u}}]{li2010making}
Mu~Li, James Tin-Yau Kwok, and Baoliang L{\"u}. 2010.
\newblock Making large-scale nystr{\"o}m approximation possible.
\newblock In \emph{Proceedings of the 27th International Conference on Machine Learning, ICML 2010}, page 631.

\bibitem[{Li et~al.(2023)Li, Allal, Zi, Muennighoff, Kocetkov, Mou, Marone, Akiki, Li, Chim et~al.}]{li2023starcoder}
Raymond Li, Loubna~Ben Allal, Yangtian Zi, Niklas Muennighoff, Denis Kocetkov, Chenghao Mou, Marc Marone, Christopher Akiki, Jia Li, Jenny Chim, et~al. 2023.
\newblock Starcoder: may the source be with you!
\newblock \emph{arXiv preprint arXiv:2305.06161}.

\bibitem[{Lialin et~al.(2023)Lialin, Muckatira, Shivagunde, and Rumshisky}]{lialin2023relora}
Vladislav Lialin, Sherin Muckatira, Namrata Shivagunde, and Anna Rumshisky. 2023.
\newblock Relora: High-rank training through low-rank updates.
\newblock In \emph{The Twelfth International Conference on Learning Representations}.

\bibitem[{Liu et~al.(2024)Liu, Wang, Yin, Molchanov, Wang, Cheng, and Chen}]{liu2024dora}
Shih-Yang Liu, Chien-Yi Wang, Hongxu Yin, Pavlo Molchanov, Yu-Chiang~Frank Wang, Kwang-Ting Cheng, and Min-Hung Chen. 2024.
\newblock Dora: Weight-decomposed low-rank adaptation.
\newblock \emph{arXiv preprint arXiv:2402.09353}.

\bibitem[{Liu(2019)}]{liu2019roberta}
Yinhan Liu. 2019.
\newblock Roberta: A robustly optimized bert pretraining approach.
\newblock \emph{arXiv preprint arXiv:1907.11692}, 364.

\bibitem[{Meng et~al.(2024)Meng, Wang, and Zhang}]{meng2024pissa}
Fanxu Meng, Zhaohui Wang, and Muhan Zhang. 2024.
\newblock Pissa: Principal singular values and singular vectors adaptation of large language models.
\newblock \emph{arXiv preprint arXiv:2404.02948}.

\bibitem[{Nemtsov et~al.(2016)Nemtsov, Averbuch, and Schclar}]{nemtsov2016matrix}
Arik Nemtsov, Amir Averbuch, and Alon Schclar. 2016.
\newblock Matrix compression using the nystr{\"o}m method.
\newblock \emph{Intelligent Data Analysis}, 20(5):997--1019.

\bibitem[{Ouyang et~al.(2022)Ouyang, Wu, Jiang, Almeida, Wainwright, Mishkin, Zhang, Agarwal, Slama, Ray et~al.}]{ouyang2022training}
Long Ouyang, Jeffrey Wu, Xu~Jiang, Diogo Almeida, Carroll Wainwright, Pamela Mishkin, Chong Zhang, Sandhini Agarwal, Katarina Slama, Alex Ray, et~al. 2022.
\newblock Training language models to follow instructions with human feedback.
\newblock \emph{Advances in neural information processing systems}, 35:27730--27744.

\bibitem[{Persson et~al.(2024)Persson, Boull{\'e}, and Kressner}]{persson2024randomized}
David Persson, Nicolas Boull{\'e}, and Daniel Kressner. 2024.
\newblock Randomized nystr$\backslash$" om approximation of non-negative self-adjoint operators.
\newblock \emph{arXiv preprint arXiv:2404.00960}.

\bibitem[{Rasooli and Tetreault(2015)}]{rasooli-tetrault-2015}
Mohammad~Sadegh Rasooli and Joel~R. Tetreault. 2015.
\newblock \href {http://arxiv.org/abs/1503.06733} {Yara parser: {A} fast and accurate dependency parser}.
\newblock \emph{Computing Research Repository}, arXiv:1503.06733.
\newblock Version 2.

\bibitem[{Si et~al.(2016)Si, Hsieh, and Dhillon}]{si2016computationally}
Si~Si, Cho-Jui Hsieh, and Inderjit Dhillon. 2016.
\newblock Computationally efficient nystr{\"o}m approximation using fast transforms.
\newblock In \emph{International conference on machine learning}, pages 2655--2663. PMLR.

\bibitem[{Touvron et~al.(2023)Touvron, Martin, Stone, Albert, Almahairi, Babaei, Bashlykov, Batra, Bhargava, Bhosale et~al.}]{touvron2023llama}
Hugo Touvron, Louis Martin, Kevin Stone, Peter Albert, Amjad Almahairi, Yasmine Babaei, Nikolay Bashlykov, Soumya Batra, Prajjwal Bhargava, Shruti Bhosale, et~al. 2023.
\newblock Llama 2: Open foundation and fine-tuned chat models.
\newblock \emph{arXiv preprint arXiv:2307.09288}.

\bibitem[{Vladymyrov and Carreira-Perpinan(2016)}]{vladymyrov2016variational}
Max Vladymyrov and Miguel Carreira-Perpinan. 2016.
\newblock The variational nystrom method for large-scale spectral problems.
\newblock In \emph{International Conference on Machine Learning}, pages 211--220. PMLR.

\bibitem[{Wang(2018)}]{wang2018glue}
Alex Wang. 2018.
\newblock Glue: A multi-task benchmark and analysis platform for natural language understanding.
\newblock \emph{arXiv preprint arXiv:1804.07461}.

\bibitem[{Wang et~al.(2024)Wang, Yu, and Li}]{wang2024lora}
Shaowen Wang, Linxi Yu, and Jian Li. 2024.
\newblock Lora-ga: Low-rank adaptation with gradient approximation.
\newblock \emph{arXiv preprint arXiv:2407.05000}.

\bibitem[{Wang et~al.(2019)Wang, Gittens, and Mahoney}]{wang2019scalable}
Shusen Wang, Alex Gittens, and Michael~W Mahoney. 2019.
\newblock Scalable kernel k-means clustering with nystrom approximation: Relative-error bounds.
\newblock \emph{Journal of Machine Learning Research}, 20(12):1--49.

\bibitem[{Wang and Zhang(2013)}]{wang2013improving}
Shusen Wang and Zhihua Zhang. 2013.
\newblock Improving cur matrix decomposition and the nystr{\"o}m approximation via adaptive sampling.
\newblock \emph{The Journal of Machine Learning Research}, 14(1):2729--2769.

\bibitem[{Williams and Seeger(2000)}]{williams2000using}
Christopher Williams and Matthias Seeger. 2000.
\newblock Using the nystr{\"o}m method to speed up kernel machines.
\newblock \emph{Advances in neural information processing systems}, 13.

\bibitem[{Xia et~al.(2024)Xia, Yu, Dang, Yang, Wu, Tian, Chang, and Lin}]{xia2024rethinking}
Tingyu Xia, Bowen Yu, Kai Dang, An~Yang, Yuan Wu, Yuan Tian, Yi~Chang, and Junyang Lin. 2024.
\newblock Rethinking data selection at scale: Random selection is almost all you need.
\newblock \emph{arXiv preprint arXiv:2410.09335}.

\bibitem[{Xiong et~al.(2021)Xiong, Zeng, Chakraborty, Tan, Fung, Li, and Singh}]{xiong2021nystromformer}
Yunyang Xiong, Zhanpeng Zeng, Rudrasis Chakraborty, Mingxing Tan, Glenn Fung, Yin Li, and Vikas Singh. 2021.
\newblock Nystr{\"o}mformer: A nystr{\"o}m-based algorithm for approximating self-attention.
\newblock In \emph{Proceedings of the AAAI Conference on Artificial Intelligence}, volume~35, pages 14138--14148.

\bibitem[{Xu et~al.(2024)Xu, Sun, Zheng, Geng, Zhao, Feng, Tao, Lin, and Jiang}]{xu2024wizardlm}
Can Xu, Qingfeng Sun, Kai Zheng, Xiubo Geng, Pu~Zhao, Jiazhan Feng, Chongyang Tao, Qingwei Lin, and Daxin Jiang. 2024.
\newblock Wizardlm: Empowering large pre-trained language models to follow complex instructions.
\newblock In \emph{The Twelfth International Conference on Learning Representations}.

\bibitem[{Yu et~al.(2023)Yu, Jiang, Shi, Yu, Liu, Zhang, Kwok, Li, Weller, and Liu}]{yu2023metamath}
Longhui Yu, Weisen Jiang, Han Shi, Jincheng Yu, Zhengying Liu, Yu~Zhang, James~T Kwok, Zhenguo Li, Adrian Weller, and Weiyang Liu. 2023.
\newblock Metamath: Bootstrap your own mathematical questions for large language models.
\newblock \emph{arXiv preprint arXiv:2309.12284}.

\bibitem[{Zhang et~al.(2023)Zhang, Chen, Bukharin, Karampatziakis, He, Cheng, Chen, and Zhao}]{zhang2023adalora}
Qingru Zhang, Minshuo Chen, Alexander Bukharin, Nikos Karampatziakis, Pengcheng He, Yu~Cheng, Weizhu Chen, and Tuo Zhao. 2023.
\newblock Adalora: Adaptive budget allocation for parameter-efficient fine-tuning.
\newblock \emph{arXiv preprint arXiv:2303.10512}.

\bibitem[{Zheng et~al.(2023)Zheng, Chiang, Sheng, Zhuang, Wu, Zhuang, Lin, Li, Li, Xing et~al.}]{zheng2023judging}
Lianmin Zheng, Wei-Lin Chiang, Ying Sheng, Siyuan Zhuang, Zhanghao Wu, Yonghao Zhuang, Zi~Lin, Zhuohan Li, Dacheng Li, Eric Xing, et~al. 2023.
\newblock Judging llm-as-a-judge with mt-bench and chatbot arena.
\newblock \emph{Advances in Neural Information Processing Systems}, 36:46595--46623.

\bibitem[{Zheng et~al.(2024)Zheng, Zhang, Shen, Liu, Lin, Fu, Chen, and Yue}]{zheng2024opencodeinterpreter}
Tianyu Zheng, Ge~Zhang, Tianhao Shen, Xueling Liu, Bill~Yuchen Lin, Jie Fu, Wenhu Chen, and Xiang Yue. 2024.
\newblock Opencodeinterpreter: Integrating code generation with execution and refinement.
\newblock \emph{arXiv preprint arXiv:2402.14658}.

\end{thebibliography}

\appendix
\section{Detailed Derivation for Nyström Approximation}
\label{app:derivation}
This section provides a detailed derivation of the Nyström approximation presented in Section \ref{sec:method}, following the approach proposed in \cite{nemtsov2016matrix}. Specifically, the quadrature technique is applied to the sample matrix of \(W\), followed by an out-of-sample extension to approximate \(W\). 

The basic quadrature technique of the Nyström method is used to approximate the Singular Value Decomposition (SVD) of a matrix. In this context, no eigen-decomposition is required. Specifically, denote the matrix \( W \in \mathbb{R}^{m \times n} \) can be decomposed as:
\begin{equation}
W = \begin{bmatrix}
A_W & B_W \\
F_W & C_W
\end{bmatrix}. \label{eq:Wsplit}  
\end{equation}
where, \( A_W \in \mathbb{R}^{r \times r} \) is designated to be the sample matrix, \( B_W \in \mathbb{R}^{r \times (n-r)} \) and \( F_W \in \mathbb{R}^{(m-r) \times r} \) represent the remaining sampled column and row components, respectively, and \( C_W \in \mathbb{R}^{(m-r) \times (n-r)} \) corresponds to the remainder of the matrix \( W \).

The derivation begins with the SVD of \( A_W \), expressed as:
\begin{equation}
A_W = U \Lambda V^T,\label{eq:uasvd}
\end{equation}
where \( U, V \in \mathbb{R}^{r \times r} \) are unitary matrices, and \( \Lambda \in \mathbb{R}^{r \times r} \) is a diagonal matrix. Assuming that zero is not a singular value of \( A_W \), the decomposition can be further approximated. Accordingly, the matrix \( U \) is formulated as:
\begin{equation}
U = A_W V \Lambda^{-1}. \label{eq:u}
\end{equation}
Let \( u^i, h^i \in \mathbb{R}^r \) represent the \( i \)-th columns of \( U \) and \( V \), respectively. Denote \( u^i = \{u^i_l\}_{l=1}^{r} \) as the individual elements of the \( i \)-th column of \( U \). Using Eq. \eqref{eq:u}, each element \( u^i_l \) is expressed as the sum: 
\begin{equation}
u^i_l = \frac{1}{\lambda_i} \sum_{j=1}^{n} W_{lj} \cdot h^i_j.
\end{equation}
The elements of \( F_W \) can be used as interpolation weights to extend the singular vector \( u^i \) to the \( k^{th} \) row of \( W \), where \( s+1 \leq k \leq n \). Let \( \tilde{u}^i = \{\tilde{u}^i_{k-s}\}_{k=s+1}^{n} \in \mathbb{R}^{n-s \times 1} \) denote a column vector comprising all the approximated entries. Each element \( \tilde{u}^i_k \) is computed as:
\begin{equation}
\tilde{u}^i_k = \frac{1}{\lambda_i} \sum_{j=1}^{n} W_{kj} \cdot h^i_j.
\end{equation}
Thus, the matrix form of \( \tilde{u}^i \) is given by \( \tilde{u}^i =\frac{1}{\lambda_i} F_W \cdot h^i \).
By arranging all the \( \tilde{u}^i \)'s into a matrix \( \tilde{U} = \left[ \tilde{u}^1 \; \tilde{u}^2 \; \dots \; \tilde{u}^r \right] \in \mathbb{R}^{n-s \times r} \), the following expression is obtained:
\begin{equation}
\tilde{U} = F_W  H\Lambda^{-1}.
\end{equation}
The Eq. \eqref{eq:uasvd} can also be written as \( V = A_W^T U \Lambda^{-1} \). To approximate the right singular vectors of the out-of-sample columns, a symmetric argument is applied, yielding:
\begin{equation}
\tilde{H} = B^T_W U \Lambda^{-1}.
\end{equation}
In that case, the full approximations of the left and right singular vectors of \( \widehat{W} \), represented by \( \tilde{U} \) and \( \tilde{H} \), respectively, are then obtained as follows:
\begin{equation}
\widehat{U} = 
\begin{bmatrix}
U \\
F_{W} V \Lambda^{-1}
\end{bmatrix},
\quad
\widehat{V} = 
\begin{bmatrix}
V \\
B_{W}^{T} U \Lambda^{-1}
\end{bmatrix}.
\end{equation}
The explicit Nyström form of \( \tilde{M} \) is given by:
\begin{align}
\widehat{W} &= \widehat{U}\Lambda \widehat{V}^{T}  \nonumber \\
&= \begin{bmatrix}
U \\
F_W V \Lambda^{-1}
\end{bmatrix}
\Lambda
\begin{bmatrix}
V^T  & \Lambda^{-1}U^TB_W
\end{bmatrix} \nonumber \\
&= \begin{bmatrix}
A_{W} & B_{W} \\
F_{W} & F_{W} A_{W}^+ B_{W}
\end{bmatrix} \nonumber \\
&= \begin{bmatrix}
A_{W}\\
F_{W}
\end{bmatrix} A_{W}^+ 
\begin{bmatrix}
A_{W} & B_{W}
\end{bmatrix},
\end{align}
where \( A_W^+ \) denotes the pseudo-inverse of \( W \). In this approximation, \( \widehat{W} \) does not modify \(A_{W}, B_{W}\) and \(F_{W}\) but approximates \( C_W \) by \( F_W  A_W^+ B_W \). This approach achieves a matrix approximation using only the selected rows and columns, effectively capturing the essential structure with reduced computational complexity.

\section{Experiments on Various Initializations}
\label{app:init}
For SLoRA, we kept the initialisation of the $A$ and $B$ matrices the same as for LoRA, and in turn explored the effect of different methods of initialisation of the intermediate matrices on the results. Specifically, we experimented with Kaiming initialization and Gaussian initialization on all the NLG tasks of LLaMA 2-7B, with the same experimental setup as in Section \ref{sec:experiments}. The performance of the models under these settings is shown in Table~\ref{table:results_initialization}. The results indicate that Kaiming initialization consistently achieves better performance across all tasks. Gaussian initialization also achieves competitive results, which demonstrates the robustness of our method. In our experiments, we use kaiming to initialize SLoRA.

\begin{table}[h]
\centering
\begin{tabular}{ccc}
\hline
\textbf{Tasks} & \textbf{Kaiming} & \textbf{Gaussian} \\ \hline
GSM8K          & \textbf{56.48}   & 56.10             \\
MATH           & \textbf{10.68}   & 9.56              \\
HumanEval      & \textbf{23.78}   & 23.2              \\
MBPP           & \textbf{42.32}   & 40.5              \\
MT-Bench       & \textbf{4.85}    & 3.93              \\ \hline
\end{tabular}
\caption{Different Initialization on SLoRA}
\label{table:results_initialization}
\end{table}

\section{Experiments on Various Learning Rates}
\label{app:lr}
We evaluated the impact of four learning rates: 2E-4, 2E-5, 5E-4 and 5E-5 on the model's performance. The experimental setup remains the same as described earlier. The results of these experiments are presented in Table~\ref{table:learning_rate_comparison}.
Among the evaluated learning rates, 5E-4 achieved the best overall performance. However, we opted for 2E-4 in our experiments, as its performance, while slightly lower than that of 5E-4, remained comparable and still exceeded the original baseline. Moreover, at the learning rate of 2E-4, NLoRA exhibited lower loss and better convergence behavior, making it a more appropriate choice for our experimental setup.

\begin{table}[]
\centering
\begin{tabular}{cccc}
\hline
\textbf{Strategy}      & \textbf{LR} & \textbf{GSM8K} & \textbf{MATH}  \\ \hline
\multirow{4}{*}{SLoRA} & 2E-4       & 56.48          & 10.68          \\
                       & 5E-4       & \textbf{59.51} & \textbf{11.04} \\
                       & 2E-5       & 51.02          & 6.94           \\
                       & 5E-5       & 52.84          & 8.36           \\ \hline
\multirow{4}{*}{NLoRA} & 2E-4       & \textbf{57.70} & 9.94           \\
                       & 5E-4       & 54.81          & \textbf{10.60} \\
                       & 2E-5       & 45.11          & 6.42           \\
                       & 5E-5       & 52.39          & 7.58           \\ \hline
\end{tabular}
\caption{Comparasion of different learning rate on SLoRA and NLoRA}
\label{table:learning_rate_comparison}
\end{table}

For the case of fine-tuning only the intermediate matrix, we tested the performance under different learning rates. The results indicate that a learning rate of 2E-3 achieved the best performance. The result is shown in \figurename~\ref{table:learning_rates_on}.

\begin{table}[]
\centering
\begin{tabular}{ccc}
\hline
\textbf{LR} & \textbf{GSM8K} & \textbf{MATH} \\ \hline
2E-4        & 43.29          & 5.74          \\
5E-4        & 44.20          & 5.70          \\
2E-3        & \textbf{44.28} & \textbf{6.86} \\
5E-3        & 40.86          & 6.08          \\ \hline
\end{tabular}
\caption{Comparasion of Different Learning Rates on IntTune}
\label{table:learning_rates_on}
\end{table}

\begin{table*}[t]
\centering
\begin{tabular}{@{}ccccccc@{}}
\toprule
\textbf{Strategy} & \textbf{Parameters} & \textbf{GSM8K} & \textbf{MATH} & \textbf{HumanEval} & \textbf{MBPP}  & \textbf{MT-Bench} \\ \midrule
LoRA              & 320M                & 42.30          & 5.50          & 18.29              & 35.34          & 4.58              \\
NLoRA             & 323M                & 57.70          & 9.94          & 25.00     & 43.12 & 4.82              \\
NLoRA+RMSProp     & 323M                & \textbf{58.10}  & \textbf{10.82}         & \textbf{25.60}              & \textbf{43.40}          & \textbf{4.99}     \\ \bottomrule
\end{tabular}
\caption{Comparision of Adamw and RMSProp on NLG}
\label{table:optim_nlg}
\end{table*}

\begin{table*}[]
\centering
\begin{tabular}{@{}ccccccccc@{}}
\toprule
\textbf{Strategy} & \textbf{MNLI}  & \textbf{SST-2} & \textbf{MRPC}  & \textbf{CoLA}  & \textbf{QNLI}  & \textbf{QQP}   & \textbf{RTE}   & \textbf{STS-B} \\ \midrule
LoRA              & 90.65          & 94.95          & 89.95          & 69.82          & 93.87          & 91.99          & 85.20          & 91.60          \\
NLoRA             & \textbf{90.74} & \textbf{96.22} & \textbf{91.91} & \textbf{73.41} & \textbf{94.45} & \textbf{92.03} & \textbf{88.09} & \textbf{92.14} \\
NLoRA+RMSProp     & 90.41          & \textbf{96.22} & \textbf{91.91} & 68.61          & 94.18          & \textbf{92.03} & \textbf{88.09} & 91.86          \\ \bottomrule
\end{tabular}
\caption{Comparision of Adamw and RMSProp on NLU}
\label{table:optim_nlu}
\end{table*}
\section{Experiments on Various Optimizers}
\label{app:optim}
We experimented with different optimizers on both NLG and NLU tasks. In addition to the default AdamW optimizer, we also evaluated the RMSProp optimizer. 
Other experimental setups are the same as Section \ref{sec:experiments}.
The experimental results are shown in \tablename~\ref{table:optim_nlg} and \tablename~\ref{table:optim_nlu}.

On NLG tasks, we observed that the RMSProp optimizer further improved the model's performance. However, its performance on NLU tasks was relatively mediocre. This discrepancy might stem from the underlying differences in the nature of NLG and NLU tasks. NLG tasks typically involve generating coherent sequences of text, which require more stable gradient updates over longer contexts. RMSProp's adaptive learning rate mechanism, which emphasizes recent gradients, may help maintain stability and enhance performance in such scenarios. In contrast, NLU tasks often involve classification or regression over shorter input sequences, where AdamW's weight decay and bias correction might be more effective in avoiding overfitting and ensuring generalization, thus outperforming RMSProp in these tasks.

\section{Experimental Settings on NLU}
\label{app:nlu}
We evaluate the performance on the GLUE benchmark, which includes two single-sentence tasks (CoLA and SST-2), three natural language inference tasks (MNLI, QNLI, and RTE), and three similarity and paraphrase tasks (MRPC, QQP, and STS-B). For evaluation metrics, we report overall accuracy (matched and mismatched) for MNLI, Matthew's correlation for CoLA, Pearson's correlation for STS-B, and accuracy for the remaining datasets.

In DeBERTa-v3-base, SLoRA and NLoRA were applied to the $W_Q$, $W_K$, and $W_V$ matrices, while in RoBERTa-large, they were applied to the $W_Q$ and $W_V$ matrices.
The experiments for natural language understanding (NLU) were conducted using the publicly available LoRA codebase. For MRPC, RTE, and STS-B tasks, we initialized RoBERTa-large with a pretrained MNLI checkpoint. The rank of SLoRA and NLoRA in these experiments was set to 8. Optimization was performed using AdamW with a cosine learning rate schedule. \tablename~\ref{table:paramnluN} and \tablename~\ref{table:paramnluS} outline the hyperparameters used for the GLUE benchmark experiments.

\begin{table*}[]
\centering
\begin{tabular}{@{}ccccc|cccc@{}}
\toprule
\multirow{2}{*}{\textbf{Dataset}} & \multicolumn{4}{c|}{\textbf{DeBERTa-v3-base}} & \multicolumn{4}{c}{\textbf{RoBERTa-large}} \\ \cmidrule(l){2-9} 
                                  & LR        & BS     & Epoch    & LoRA alpha    & LR       & BS    & Epoch    & LoRA alpha   \\ \midrule
CoLA                              & 3E-04     & 16     & 40       & 16            & 4E-04    & 8     & 20       & 8            \\
SST-2                             & 5E-04     & 16     & 10       & 8             & 5E-04    & 16    & 10       & 8            \\
MRPC                              & 5E-04     & 32     & 100      & 16            & 2E-04    & 32    & 50       & 16           \\
MNLI                              & 3E-04     & 32     & 10       & 16            & 3E-04    & 32    & 10       & 16           \\
QNLI                              & 2E-04     & 32     & 20       & 16            & 6E-04    & 16    & 10       & 8            \\
QQP                               & 6E-04     & 32     & 20       & 8             & 6E-04    & 16    & 10       & 16           \\
RTE                               & 3E-04     & 32     & 40       & 16            & 5E-04    & 32    & 30       & 16           \\
STS-B                             & 5E-04     & 16     & 10       & 16            & 3E-04    & 16    & 30       & 16           \\ \bottomrule
\end{tabular}
\caption{Hyperparameters of NLoRA on GLUE}
\label{table:paramnluN}
\end{table*}

\begin{table*}[]
\centering
\begin{tabular}{@{}ccccc|cccc@{}}
\toprule
\multirow{2}{*}{\textbf{Dataset}} & \multicolumn{4}{c|}{\textbf{DeBERTa-v3-base}} & \multicolumn{4}{c}{\textbf{RoBERTa-large}} \\ \cmidrule(l){2-9} 
                                  & LR        & BS     & Epoch    & LoRA alpha    & LR       & BS    & Epoch    & LoRA alpha   \\ \midrule
CoLA                              & 3E-04     & 16     & 40       & 16            & 4E-04    & 8     & 20       & 8            \\
SST-2                             & 5E-04     & 16     & 10       & 8             & 5E-04    & 16    & 10       & 8            \\
MRPC                              & 5E-04     & 32     & 100      & 16            & 2E-04    & 32    & 50       & 16           \\
MNLI                              & 3E-04     & 32     & 10       & 16            & 3E-04    & 32    & 20       & 16           \\
QNLI                              & 2E-04     & 32     & 20       & 16            & 6E-04    & 16    & 10       & 8            \\
QQP                               & 6E-04     & 32     & 20       & 8             & 6E-04    & 16    & 10       & 16           \\
RTE                               & 3E-04     & 32     & 40       & 16            & 5E-04    & 32    & 30       & 16           \\
STS-B                             & 5E-04     & 16     & 10       & 16            & 3E-04    & 16    & 30       & 16           \\ \bottomrule
\end{tabular}
\caption{Hyperparameters of SLoRA on GLUE}
\label{table:paramnluS}
\end{table*}

For IntTune, we set both the LoRA rank and LoRA alpha to 8. The remaining parameter configurations are provided in \tablename~\ref{table:deberta_on_param}.

\begin{table*}[]
\centering
\begin{tabular}{cccc}
\hline
\textbf{Dataset} & LR    & BS & Epoch \\ \hline
CoLA             & 7E-03 & 16 & 40    \\
SST-2            & 6E-03 & 32 & 30    \\
MRPC             & 4E-03 & 16 & 50    \\
MNLI             & 6E-03 & 64 & 20    \\
QNLI             & 8E-03 & 64 & 20    \\
QQP              & 6E-03 & 32 & 20    \\
RTE              & 6E-03 & 16 & 25    \\
STS-B            & 6E-03 & 16 & 60    \\ \hline
\end{tabular}
\caption{Hyperparameters for IntTune}
\label{table:deberta_on_param}
\end{table*}

\end{document}